\documentclass[twocolumn]{article}

\usepackage[width=17cm,height=22cm]{geometry}
\usepackage[utf8]{inputenc}
\usepackage{fancyvrb}
\usepackage{authblk}
\usepackage{hyperref}
\usepackage{graphicx}
\usepackage{amsmath}

\usepackage{tabularx}
\usepackage{amssymb}
\usepackage{pifont}
\newcommand{\cmark}{\ding{51}}%

\title{Feature extraction with regularized siamese networks for outlier detection: application to lesion screening in medical imaging}
\author{Z. ALAVERDYAN} 
\author{C. LARTIZIEN}
\affil{Univ Lyon, INSA‐Lyon, Université Claude Bernard Lyon 1, UJM-Saint Etienne, CNRS, Inserm, CREATIS UMR 5220, U1206, F‐69621, LYON, France}
\date{June 6, 2017}

\begin{document}
\maketitle

\begin{abstract}
Computer aided diagnosis (CAD) systems are designed to assist clinicians in various tasks, including highlighting abnormal regions in a medical image. A common approach consists in training a voxel-level binary classifier on a set of feature vectors extracted from normal and pathological areas in patients' scans. However, many pathologies (such as epilepsy) are characterized by lesions that may be located anywhere in the brain, have various shapes, sizes and texture. An adequate representation of such a heterogeneity requires a significant amount of annotated data which is a major issue in the medical domain. Therefore, we built on a previously proposed approach that considers epilepsy lesion detection task as a voxel-level outlier detection problem. It consists in building a oc-SVM classifier for each voxel in the brain volume using a small number of clinically-guided features \cite{azami2016}.\\
Our goal in this study is to make a step forward by replacing the handcrafted features with automatically learnt representations using neural networks. We propose a novel version of siamese networks trained on patches extracted from healthy patients' scans only. This network, composed of stacked autoencoders as subnetworks, is regularized by the reconstruction error of the patches. It is designed to learn representations that bring patches centered at the same voxel localization 'closer' with respect to the chosen metric (i.e. cosine). Finally, the middle layer representations of the subnetworks are fed to oc-SVM classifiers at voxel-level. The method is validated on 3 patients' MRI scans with confirmed epilepsy lesions and shows a promising performance.
\end{abstract}

\medskip


\section{Introduction}

Computer aided diagnosis (CAD) systems have been introduced as to assist clinicians in various tasks such as tumor segmentation, detection of abnormal regions in a medical image, etc. Recent CAD systems exploit various modalities of neuroimaging data, such as magnetic resonance imaging (MRI) and positron emission tomography (PET).
A frequently applied approach in the existing CAD systems \cite{hong2014,zhao2016} assumes extracting voxel-level descriptors and feeding them to a classification algorithm that would learn how to separate the suspicious voxels from the healthy ones. However, such approaches are hard to exploit when the number of pathological cases in the training set is not sufficient to account for the complexity of the task. In  particular, epilepsy lesions vary largely in terms of shapes, sizes, textures and it is not trivial to obtain a well-annotated dataset to represent such a variability.   
Therefore, in \cite{azami2016} we proposed to adapt a different approach which consists in treating the epilepsy lesion detection in brain magnetic resonance (MR)  images as an outlier detection problem. For each voxel in the brain volume, six clinically-guided features were extracted and fed into a oc-SVM classifier. These oc-SVM models (one per voxel) were trained solely on the voxels belonging to healthy brain scans and  allowed detecting epilepsy lesions as outliers when tested on MR scans carrying pathologies.\\
This work extends the previous approach by replacing the handcrafted features with automatically extracted representations learned with a siamese network. The network is composed of stacked denoising autoencoders and is trained on the patches of healthy brain volumes only, by utilizing a novel loss function adapted to the given context. We expect such a network to be efficient in producing useful representations in our outlier detection context.\\

\section{Method}
In this work we make an attempt to learn \textit{patch-level representations} in an unsupervised manner and use them as feature vectors for oc-SVM classifiers at voxel-level. The representations are learnt on a set of patches extracted from MRI scans of healthy patients only, hence pathological examples are not exploited. Our approach could be beneficial in the settings where it is not feasible to  sample an adequate amount of pathological cases. 
\\
We aim at finding a mapping from the original patches to a representation space where the patches centered at the same voxel but belonging to different patients (such patches will be referred to as "similar") are "close". Considering each set of similar patches as representatives of a certain class, we would have as many classes as there are voxels in a typical brain volume (about 4 millions), while having a far smaller number of examples per class. Siamese networks have been successfully applied in learning such a mapping when the number of classes is much larger than the number of samples in each of them [\cite{bromley1993,chopra2005}].    
\begin{figure}
    \centering
    \includegraphics[width=0.5\textwidth]{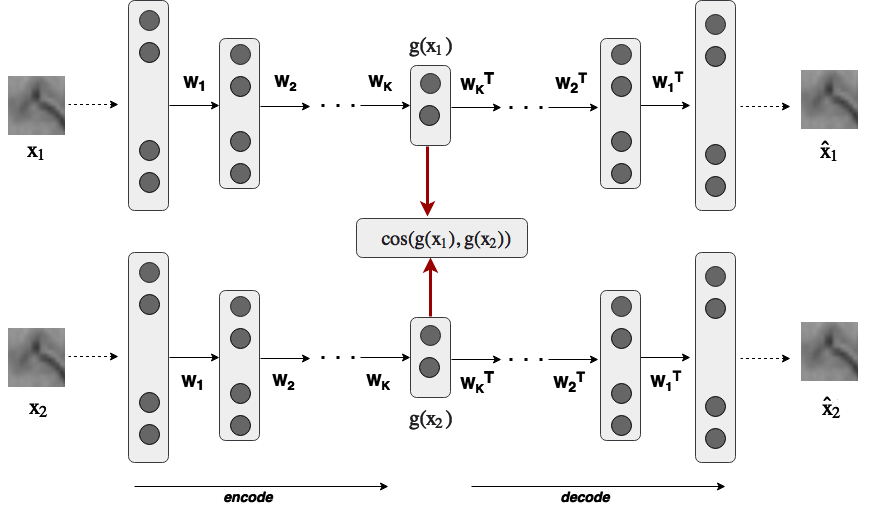}
    \caption{Siamese neural network with stacked autoencoders with tied weights.}
    \label{RSN}
  \end{figure}

\subsection{Regularized siamese architecture for feature extraction}
\subsubsection{Architecture}
The proposed architecture is illustrated on figure \ref{RSN}. Our regularized siamese neural network (rSNN) consists of two identical (same architecture, shared parameters) subnetworks -  stacked denoising autoencoders (sDA) with $K$ hidden layers and a cost module. Each layer $k$ of the sDAs is associated with a weight matrix $W_{k}$, that connects the neurons in the layer $k-1$ to those in the $k$-th layer, and a bias vector $\mathbf{b_{k}}$. A denoising autoencoder \cite{vincent2008} receives a distorted version $ \mathbf{\tilde{x}_i}$ of a patch $\mathbf{x_i}$ at input  and yields a reconstructed output $\mathbf{\hat{x}_i}$. The parameters are iteratively updated to optimize a loss function that measures the deviation between $\mathbf{\hat{x}_i}$ and the clean input $\mathbf{x_i}$. The siamese network receives a pair of patches $(\mathbf{x_{1}}, \mathbf{x_{2}})$ at input, then each patch is propagated through the corresponding subnetwork yielding representations $g(\mathbf{x_{t}}), t = (1,2)$ in the middle (narrow) layer which are then passed to the loss function $L$ below. Unlike in the classical siamese frameworks where the network also receives a binary label that stands for the similarity/dissimilarity of the pair, in our application all the considered pairs are similar and therefore the label is not present in the loss function. The loss function, however, can be easily modified to meet the typical setting. Another version of regularized siamese networks has been proposed by \cite{chen2011}, however, within the classical context.

\subsubsection{Loss function}

Our loss function is designed to maximize the cosine similarity between $g(\mathbf{x_{1}})$ and $g(\mathbf{x_{2}})$. In the absence of dissimilar pairs (the notion of dissimilar patches is not defined in our context) it is necessary to add a regularizing term. To this end we propose to use the mean squared error between the input patches and their reconstructions output by the subnetworks. Without a proper regularization term the loss function could be driven to 0 by mapping all the patches to a constant value. The proposed loss function for a single pair hence is:\\
\begin{equation}
\label{loss}
L(\mathbf{x_1}, \mathbf{x_2}; \Theta) = \alpha \sum_{t=1}^{2} { {||\mathbf{x_{t}} -\mathbf{\hat{x}_{t}}||_2^2}} - (1-\alpha)  {cos(g(\mathbf{x_{1}}), g(\mathbf{x_{2}}))}
\end{equation}
$ $\\
where $\mathbf{\hat{x}_{t}}$ is the reconstructed output of subnetwork $t$ of the patch $\mathbf{x_{t}}$ while $g(\mathbf{x_{t}})$ is its representation in the middle layer and $\alpha$ is a coefficient that controls the tradeoff between the two terms. $\Theta$ represents the parameter set. (Note that in our case the input is scaled between 0 and 1).
\subsubsection{Training}
The training of the network is achieved in two steps. In the first step, one of the stacked denoising autoencoder-subnetworks is pre-trained using greedy layer-wise pretraining \cite{bengio2009}. This allows to initialize the network parameters for the next fine-tuning step. Since the subnetworks are identical the parameter values are copied to initialize the other subnetwork.
In the fine-tuning step the network parameters are iteratively updated as to minimize the loss function (\ref{loss}) following a stochastic gradient descent method. Each parameter is updated with the sum of the gradients of the two subnetworks with respect to the corresponding parameter.


\subsection{Voxel-level outlier detection with oc-SVM classifiers}
A oc-SVM \cite{scholkopf2001} classifier seeks to find the optimal hyperplane that separates the given points from the origin in a dot product space defined by some kernel function $\phi$. The optimization problem to be solved is the following:

\begin{equation*}
\begin{aligned}
& \underset{w, \rho, \xi_i}{\text{min} }
& & \mathrm{ \frac{1}{2} ||w||^2 + \frac{1}{\nu n} \sum_{i=1}^{n} \xi_i - \rho } \\
& \text{subject to}
& &  \mathrm{ w \cdot \phi(x_i) \geq \rho - \xi_i, \xi_i \geq 0,  i \in [1, n] }
\end{aligned}
\end{equation*}
where $n$ is the number of training examples, $\mathrm{x_i}$ is the $i$-th example in the dataset $X$, $\xi_i$-s are slack variables relaxing the inequality constraints as to account for the non-separable classes,  $\mathrm{w}$ and $\rho$ define the separating hyperplane, $\nu$ is a parameter that sets a boundary to the fraction of outliers allowed. The decision function, then, for an example $\mathrm{x}$ is $\mathrm{ w \cdot \phi(x) - \rho}$. This decision function contributes to the signed score output by a oc-SVM model (in a typical scenario examples with negatives scores would be considered outliers). \\
To validate the usefulness of the features learnt by the proposed method we use the \textit{representations in the middle layer of the subnetworks} to train oc-SVM classifiers at voxel level. Each voxel is associated with a classifier, hence the number of classifiers is equal to the number of voxels in a volume (around 4 million voxels). For a given voxel $v_i$, the associated oc-SVM classifier $C_i$ is trained on the matrix $M_i = [\mathbf{x_{i1}}, ...,  \mathbf{x_{in}}]$ where $\mathbf{x_{ij}}$ is the feature vector corresponding to the patch centered at $v_i$ of patient $j$ and $n$ is the number of patients. The length of $\mathbf{x_{ij}}$ is equal to the number of neurons in the middle layer. \\
For a new patient, each voxel $v_i$ is matched against the corresponding classifier $C_i$ and is assigned the signed score output by the classifier. This yields a "distance" map for the given patient. This map is later thresholded for each patient individually (the threshold is chosen as the score corresponding to a pre-chosen $p$-value in the distribution of scores of the patient) and 26-connectivity rule is applied. Further, the voxel clusters smaller than a fixed size (set to 82 voxels in this study) are discarded.
\section{Experiments and results}

\subsection{Dataset description and pre-processing}
Our database consists of T1-weighted MR images of 96 healthy subjects (T1 weighted MRI acquisition is a standard exam in a clinical routine where contrast in the image is due to the different relaxation time-properties of the tissues). 29 of those (\textit{DB1}) had a 3D anatomical T1-weighted brain MRI sequence (TR/TE 9.7/4 ms; 176 sagittal slices of 256 x 256 millimetric cubic voxels) on a 1.5 T Sonata scanner (Siemens Healthcare, Erlangen, Germany). The remaining 67 (\textit{DB2}) had a 3D anatomical T1-weighted brain MRI sequence on the same scanner but with a slightly different protocol (TR/TE 2400/3.55; 160 sagittal slices of 192 x 192 1.2mm cubic voxels). All the volumes were normalized to the standard brain template of the Montreal Neurological Institute (MNI) \cite{mazziotta2001} using a voxel size of 1 x 1 x 1 mm. This step is important in assuring the voxel-level correspondance between the subjects. We validated the method on 3 patients with confirmed epileptogenic lesions - all 3 patients acquired with the same scanner and the same parameters as the subjects from \textit{DB1} and having a positive MRI screening ($MRI^+$) meaning that the lesion was visually detected on the MR scan and eventually outlined by a neurologist.
  \begin{figure*}
    \centering
    \includegraphics[width=0.75\textwidth]{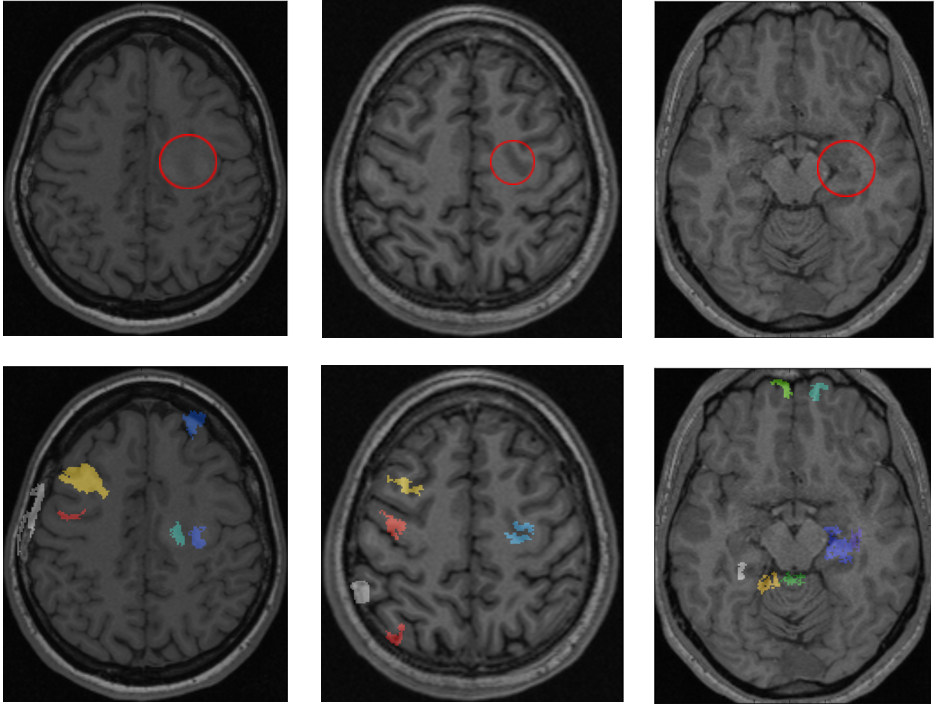}
    \caption{First row: pathological slices of patient A, B and C respectively. The true lesions are outlined in red circles. Second row: Maximum Intensity Projections of the cluster maps onto the same slice as in the first row. The maps are reported for a \textit{p-value} = 0.003. Note that some clusters appear outside of the brain volume as a result of the MIP projection. Note also that when MIP is used, multiple clusters can appear jointly on the projection, hence the number of false positive clusters on the visualization can be smaller than the reported number in the table below.}
    \label{breal}
  \end{figure*}
\subsection{Feature extraction with rSNN}
The results below are reported for the best network among the ones we tried (the networks were designed by varying the essential parameters and configurations such as the input patch size, the corruption rate, the parameter $\alpha$). The proposed rSNN subnetworks are stacked denoising autoencoders with 2 hidden layers consisting of 64 and 32 neurons respectively. 
During the pretraining the distortion rate (the probability of masking a voxel in a patch i.e. setting it to 0) was set to 0.3 and 0.1 for the first and second layers respectively. The same rate in the fine-tuning step was set to 0.1.
We extracted 9 x 9 patches with stride 5 from all available volumes of the healthy controls, which gave around 14 million patches in total, and used them in pre-training. The same number of similar pairs was used to fine-tune the model.

\subsection{oc-SVM classifier design}
We used oc-SVM classifiers with RBF kernel which gives us two parameters to set - $\nu$ (upper bound on the fraction of permitted outliers) and $\gamma$ (the kernel parameter). We empirically set the parameter $ \nu$ to 0.03 for all the oc-SVMs. The RBF kernel width, $\gamma$ was set for each voxel individually, using the median of the pairwise distances between the points of the corresponding matrix as it was suggested in \cite{caputo2002appearance}. 

\subsection{Experimental results}

Below we report the results obtained with the best model among the tried ones - rSNN trained on 9 x 9 patches with $\alpha=0.66$ (we varied the size of the input patches and the parameters $\alpha$ ). We also evaluated the performance of the same pipeline when using only one of the subnetworks i.e. a stacked denoising autoencoder with 2 hidden layers (will be referred to as sDA). To evaluate the performance of the considered models we follow the following steps. As explained in section 2.2, for each voxel localization in the brain a oc-SVM model is trained on the voxels of the healthy subjects extracted from the corresponding localization. When given a test patient's MR image, the CAD system, for each voxel of the volume, gathers the score/output of the corresponding oc-SVM model. This results in a volume of the same dimensions as the original MR image, where each voxel now is the signed score assigned by the oc-SVM model (we refer to this output as \textit{distance map}). We then calculate the score distribution of the patient and find the score corresponding to the a fixed $p$-value. The voxels with score above this value are discarded. The remaining voxels are checked against the 26-connectivity rule which produces connected components (we will refer to those as clusters). Clusters containing less than 82 voxels are discarded and the remaining constitute the final \textit{cluster map}. \\Table \ref{results_tab} reports the true lesion detections/clusters and the number of false positive clusters(clusters that were detected by the system but were not considered epileptogenic by the neurologist are considered false positive in this application). The results show that both rSNN and sDA perform adequately and manage to detect the lesions. rSNN, however, outputs less irrelevant clusters (only one epileptogenic lesion was pointed out per patient by the neurologist).
An important step of evaluation is to visualize the output cluster maps. Figure \ref{breal} demonstrates the maximum intensity projections of the detected clusters onto 3 slices of interest for patients A, B and C. The true lesion locations are outlined in red circles. The maps were obtained for a \textit{p-value} = 0.003, a threshold which allowed a clear detection of the lesions. This threshold can be varied by a physician allowing to find anomalies on different scales. As we can see, the true lesions (the ground truth is highlighted in red circles) are well detected while some of the false detections (at the brain and skull interface) can be easily eliminated by a trained eye or by post-processing the cluster maps based on geometric features. Some reported clusters may also correspond to true anomalies that are either benign or were not reported as epileptogenic by the neurologist. 




\setlength{\tabcolsep}{0.25em}
\begin{table}
\centering
\caption{The results obtained with sDA and rSNN (with $\alpha$ =0.66) reported for $p$-value = 0.003}
\label{results_tab}
\begin{tabular}{l|c|c|c}
\hline
\textbf{Model}  & \textbf{patient $A$} & \textbf{patient $B$}& \textbf{patient $C$}  \\
\hline
sDA  & \cmark(9) &\cmark(7)	&	\cmark(8)	\\     
rSNN / $ \alpha$ =0.66 & \cmark(5)   & \cmark(4)     & \cmark(7)  \\ 
\hline
\end{tabular}
\end{table}

\section{Conclusion}
This work presents an approach of feature extraction by the means of a regularized siamese network in the context of voxel-level outlier detection task. The preliminary experiments showed a promising performance when applying the method to the epilepsy lesion detection in T1-weighted MRI scans. The performance of the learnt representations could be improved by replacing the stacked denoising autoencoder-subnetworks with their convolutional counterparts. Additionally, other metrics could be explored to replace the cosine similarity in the proposed loss function.
Regarding the specific application of epilepsy lesion detection, the proposed method would gain in efficiency if other medical imaging modalities, such as PET and/or FLAIR, were considered. Incorporating multiple modalities in a single model is one of the perspectives of the future work.

\section*{Acknowledgements}
\thanks{We thank Dr Julien Jung from the Lyon Neurological Hospital for providing the MRI data set and for sharing his clinical expertise on the epilepsy patients.
}

\bibliography{cap2017}

\end{document}